\documentclass[journal]{IEEEtran}
\usepackage{float}
\usepackage[T1]{fontenc} 
\usepackage{cite}
\usepackage{amsmath,amssymb,amsfonts}
\usepackage{multirow}
\usepackage{algorithmic}
\usepackage[ruled,linesnumbered]{algorithm2e}
\usepackage{graphicx}
\usepackage{textcomp}
\usepackage{xcolor}
\usepackage{subfigure}
\usepackage{threeparttable}
\usepackage{booktabs}
\usepackage{svg}
\usepackage[numbers,sort&compress]{natbib}
\usepackage{caption}

\ifCLASSINFOpdf
\else
\fi

\begin{document}
%
\title{Discrete States-Based Trajectory Planning for Nonholonomic Robots}
%
%
%

\author{Ziyi~Zou,
        Ziang~Zhang,
        Zhen~Lu,
        Xiang~Li,
        You~Wang,
        Jie Hao,
        and Guang~Li
\thanks{\textit{Corrsponding author: You Wang}.}
\thanks{Ziyi Zou, Ziang Zhang, Zhen lu, Xiang Li, You Wang and Guang Li are with the
State Key Laboratory of Industrial Control Technology,
Institute of Cyber Systems and Control, 
Zhejiang University, Hangzhou, 310027, China
(e-mail: ziyi.Zou@zju.edu.cn, zhangziang@zju.edu.cn, zhenlulz@zju.edu.cn, 22132029@zju.edu.cn,
king\_wy@zju.edu.cn, guangli@zju.edu.cn).
}
\thanks{Jie Hao is with the Luoteng Hangzhou Techonlogy Co., Ltd., Hangzhou
310027, China (e-mail: mac@rotunbot.com).}
}

%
%


\maketitle

\begin{abstract}
Due to nonholonomic dynamics, the motion planning of nonholonomic robots is always a difficult problem. This letter presents a Discrete States-based Trajectory Planning(DSTP) algorithm for autonomous nonholonomic robots. The proposed algorithm represents the trajectory as x and y positions, orientation angle, longitude velocity and acceleration, angular velocity, and time intervals. More variables make the expression of optimization and constraints simpler, reduce the error caused by too many approximations, and also handle the gear shifting situation. L-BFGS-B is used to deal with the optimization of many variables and box constraints, thus speeding up the problem solving. Various simulation experiments compared with prior works have validated that our algorithm has an order-of-magnitude efficiency advantage and can generate a smoother trajectory with a high speed and low control effort. Besides, real-world experiments are also conducted to verify the feasibility of our algorithm in real scenes. We will release our codes as ros packages.
\end{abstract}

\begin{IEEEkeywords}
Trajectory Optimization, motion and path planning, nonholonomic motion planning
\end{IEEEkeywords}

\IEEEpeerreviewmaketitle

\section{Introduction}

\IEEEPARstart{A}{utonomous} moving robots are used in a wide range of scenarios due to their flexibility and reliability. And motion planning is a fundamental part of robots. The goal of motion planning for robots is to generate a smooth trajectory that makes the robot reach the end state safely as soon as possible. Meanwhile, in the real world, motion planning always requires fast updates to cope with changing environments and inaccuracies in localization. However, the motion planning of nonholonomic robots faces the following problems:
1) \textit{Nonholonomic dynamic constraints} impose a strong non-convexity on the planning problem, making the optimization difficult.
2) \textit{Precise obstacle collision avoidance} requires the trajectory to ensure safety and
pass through narrow areas.
3) \textit{Efficiency and quality} are both important for motion planning, but we often need to make a tradeoff between them.
4) \textit{Time information} is an inherent attribute of trajectories. However, path/speed coupled optimization increases the complexity of the problem, and optimizing separately makes it difficult to obtain optimal solutions.

\begin{figure}[t]
    \centering
        \includegraphics[width=8.5cm]{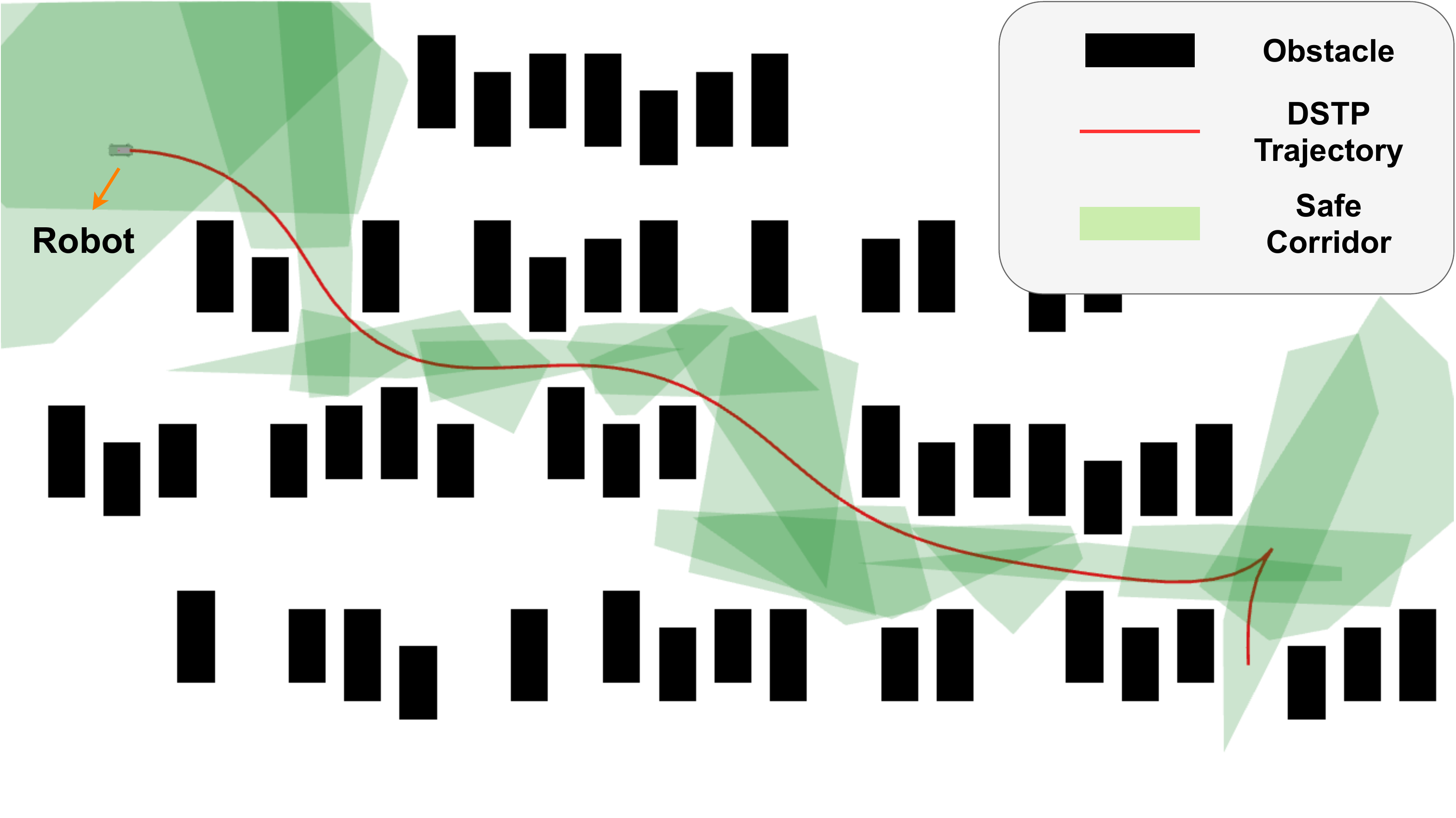}
    \caption{Illustration of trajectory generated by DSTP. \label{fig:trajectory}}
\end{figure}

Many algorithms based on discrete states, in order to reduce the number of variables and speed up the problem solving, often use fewer states and more approximations for trajectory planning\cite{TEB2,CES}, which usually increases the complexity of the expression and the error of solutions.
In this paper, we propose a novel efficient \textbf{D}iscrete \textbf{S}tates-based \textbf{T}rajectory \textbf{P}lanning method called \textbf{DSTP}, that addresses the above challenges and can generate high-quality trajectories with nonholonomic constraints and handle gear shifting situation as shown in Fig.\ref{fig:trajectory}. Besides, DSTP can significantly reduce the time required for subsequent replannings by using the previous optimization solution to warm start the solving.
Firstly, DSTP will receive the rough path planned by the front-end, a lightweight Hybrid A*\cite{dolgov2010path} planner, and then generates an x-y 2-dimension safe corridor composed of several convex polygons along the initial path using the corridor construction algorithm in \cite{liu2017planning}. Two circles are used to cover our robot as the collision model. Then we take the robot's position, orientation angle, longitudinal velocity and acceleration, angular velocity of each point, and the time interval between two states as the optimization variables. Then we formulate the trajectory planning as a nonconvex optimization problem, and use L-BFGS-B\cite{L-BFGS-B} to handle the box constraints of velocities, accelerations , and time intervals. In addition, we use trigonometric functions instead of angle constraints to handle abrupt angle changes(such as from $-\pi$ to $\pi$). After the first trajectory planning, we use the solution obtained from the previous optimization as the initial guess of the current optimization to greatly speed up the solutions.
Simulation benchmark comparisons with DL-IAPS+PJSO\cite{PJSO} and TEB\cite{TEB2} in three different scenarios are conducted to demonstrate the superiority of our algorithm. we also perform real-world tests with a car-like robot in a garage to validate our algorithm on real platforms. We summarize our contributions as:

1) We represent the robot's trajectory with discrete states, control inputs, and time intervals, which makes the trajectory optimization problem simpler and reduce the error caused by too many approximations.

2) We propose a novel and real-time discrete states-based trajectory planning method for nonholonomic robots, which can generate high-quality trajectories with nonholonomic constraints and handle gear shifts.

3) We integrate the trajectory planning method into an autonomous system, and open source the code, aiming to facilitate future progress in the field of motion planning for autonomous robots\footnote{https://github.com/ziyi-joe/DSTP}.

\section{Related Work}
The algorithms for trajectory optimization can be classified into path/speed coupled optimization and decoupled optimization.
\subsection{Path/speed Coupled Optimization}
Path/speed coupled optimization methods can get better trajectories, but it also increases the computational complexity. Zhang et al.\cite{OBCA} propose the optimization-based collision avoidance(OBCA) algorithm, assuming the obstacles are convex and reformulating the safety constraints and auxiliary decision variables as a set of smooth constraints. However, the quality of the solutions is strongly dependent on the initial guess. To provide an initial solution to help the solving part, Zhang et al.\cite{H-OBCA} propose a hierarchical algorithm, H-OBCA, using the Hybrid A* to compute a coarse path to warm-start OBCA.
The timed elastic band(TEB)\cite{TEB1} introduces temporal information based on the elastic band\cite{EB, quinlan1995real}, taking geometric and dynamic constraints into account. Now it is extended to car-like robots\cite{TEB2} and egocentric local planning representation\cite{egoTEB}.
Li et al.\cite{li2021optimization} use two circles to cover the vehicle body. Then the vehicle shrinks to two points by dilating the obstacles. Besides, trigonometric functions instead of angle constraints are used to handle angle changes in free space.
Han et al.\cite{han2022differential} propose a differential flatness-based trajectory planning and guarantee safety by confining the full vehicle model to the corresponding convex safe polygon.

\subsection{Path/speed Decoupled Optimization}
The Convex Elastic Smoothing(CES)\cite{CES} divides the trajectory optimization problem into path optimization and time optimization when fixing the other. The curvature constraint is converted into a quadratic constraint by approximating that the length of the smoothed path is roughly equal to the input initial path. However, as shown in \cite{PJSO}, the maximum curvature constraints in CES tend to be invalid when the smoothed path is much shorter. Based on CES, Zhou et al.\cite{PJSO} propose dual-loop iterative anchoring path smoothing(DL-IAPS) and piecewise-jerk speed optimization(PJSO). DL-IAPS performs path smoothing through inner and outer loops, with the inner for curvature constrained path smoothing solved by SCP and the outer for collision avoidance. PJSO divides the entire trajectory time into segments with the same time interval and assumes the jerk is constant in each segment. Then the speed optimization is formulated as a quadratic problem and solved by OSQP\cite{OSQP}.
EGO-Planner\cite{EGO} first generates the trajectory with a gradient-based spline optimizer, which formulates the problem as unconstrained nonlinear optimization with smoothness, collision, and feasibility penalty terms. Then the post-refinement procedure will reallocate the time if the trajectory violates dynamical limits.

\section{Discrete States of Nonholonomic Robots}\label{sec:Discrete_States}
\subsection{Problem Statement}\label{problem_statement}
In this paper, the trajectory is represented by a series of discrete states. At step $k$ of the trajectory, the ego robot's state can be described with position $\mathbf{p}_k=[x_k,y_k]^T$, orientation $\theta_k$, longitudinal velocity $v_k$, longitudinal acceleration $a_k$, and angular velocity $\omega_k$. They can be divided into $\mathbf{s}_k=[x_k,y_k,\theta_k,v_k]$ and control inputs $\mathbf{c}_k=[a_k,\omega_k]$. In addition, the time interval between $\mathbf{s}_k$ and $\mathbf{s}_{k+1}$ is defined as $t_k$ and we assume the jerk and angular acceleration are constant from $t_k$ to $t_{k+1}$

Another thing to note is the orientation angle of robot, which we often normalize to $(-\pi,\pi]$. However, the orientation sometimes changes through $-\pi$ to $\pi$(or vice versa), leading to abrupt angle change. In this paper, we use trigonometric function to express the equality of two angle. Here we define:
\begin{equation}
    \begin{aligned}
        a \triangleq b &\Longleftrightarrow \begin{cases}
            \sin{a}=\sin{b} \\ \cos{a}=\cos{b}
        \end{cases}\\
        \mathbf{s}_1 \triangleq \mathbf{s}_2 &\Longleftrightarrow \begin{cases}
            x_1=x_2 \\ y_1=y_2 \\ \theta_1 \triangleq \theta_2 \\ v_1=v_2
        \end{cases}
    \end{aligned}
\end{equation}

There are many kinematic and dynamical constraints nonholonomic robots. In the remainder of this section, we will introduce these constraints.

\subsection{Waypoints Constraints}
The robot needs to pass through certain waypoints in the work, such as the start and target positions. Usually, The robot needs to reach specified states at these path points. For the start and the goal, we have:
\begin{equation}
    \begin{aligned}
        \mathbf{s}_0&\triangleq \mathbf{s}_s\\
        \mathbf{s}_n&\triangleq \mathbf{s}_g,\\
    \end{aligned}
\end{equation}
where $\mathbf{s}_s$ and $\mathbf{s}_g$ are the start and goal states respectively.

\subsection{Dynamic Feasibility}
\subsubsection{Redundant States Relationship}
The states in \ref{problem_statement} are redundant, which indicates that they are not mutually independent. The purpose of adding redundant states is to simplify the optimization problem as we can explicitly write out certain states such as longitude velocity, acceleration and angular velocity of the robot. We can write the relationship between them according to the dynamics as:
\begin{equation}\label{eq:relationship}
    \begin{aligned}
        x_{k+1} &\approx x_k + \frac{v_kcos\theta_k+v_{k+1}cos\theta_{k+1}}{2} t_k\\
        y_{k+1} &\approx y_k + \frac{v_ksin\theta_k+v_{k+1}sin\theta_{k+1}}{2} t_k\\
        v_{k+1} &= v_k + a_kt_k + \frac{1}{2}\frac{a_{k+1}-a_k}{t_k}t_k^2\\
                &= v_k + \frac{a_{k+1}-a_k}{2}t_k\\
        \theta_{k+1} &\triangleq \theta_k + \omega_kt_k + \frac{1}{2}\frac{\omega_{k+1}-\omega_k}{t_k}t_k^2\\
        &\triangleq \theta_k + \frac{\omega_{k+1}+\omega_k}{2}t_k\\
        \forall k\in&\{0,1,2,...,n-1\}.
    \end{aligned}
\end{equation}

\subsubsection{Velocity and Acceleration Limit}\label{sec:limited_v_a}
The longitude velocity and acceleration for autonomous robots are always limited within a reasonable range by physical factors. They can usually be written simply as the box constrains:
\begin{equation}\label{eq:limited_v_w}
    \begin{aligned}
        v_{min} &\leq v_k \leq v_{max}\\
        a_{min} &\leq a_k \leq a_{max}\\
        \forall k\in&\{0,1,2,...,n\}.
    \end{aligned}
\end{equation}

The box constraint ares quite simple, which limits only the upper and lower bounds of variables.

\begin{figure}[t]
    \centering
        \includegraphics[width=8.5cm]{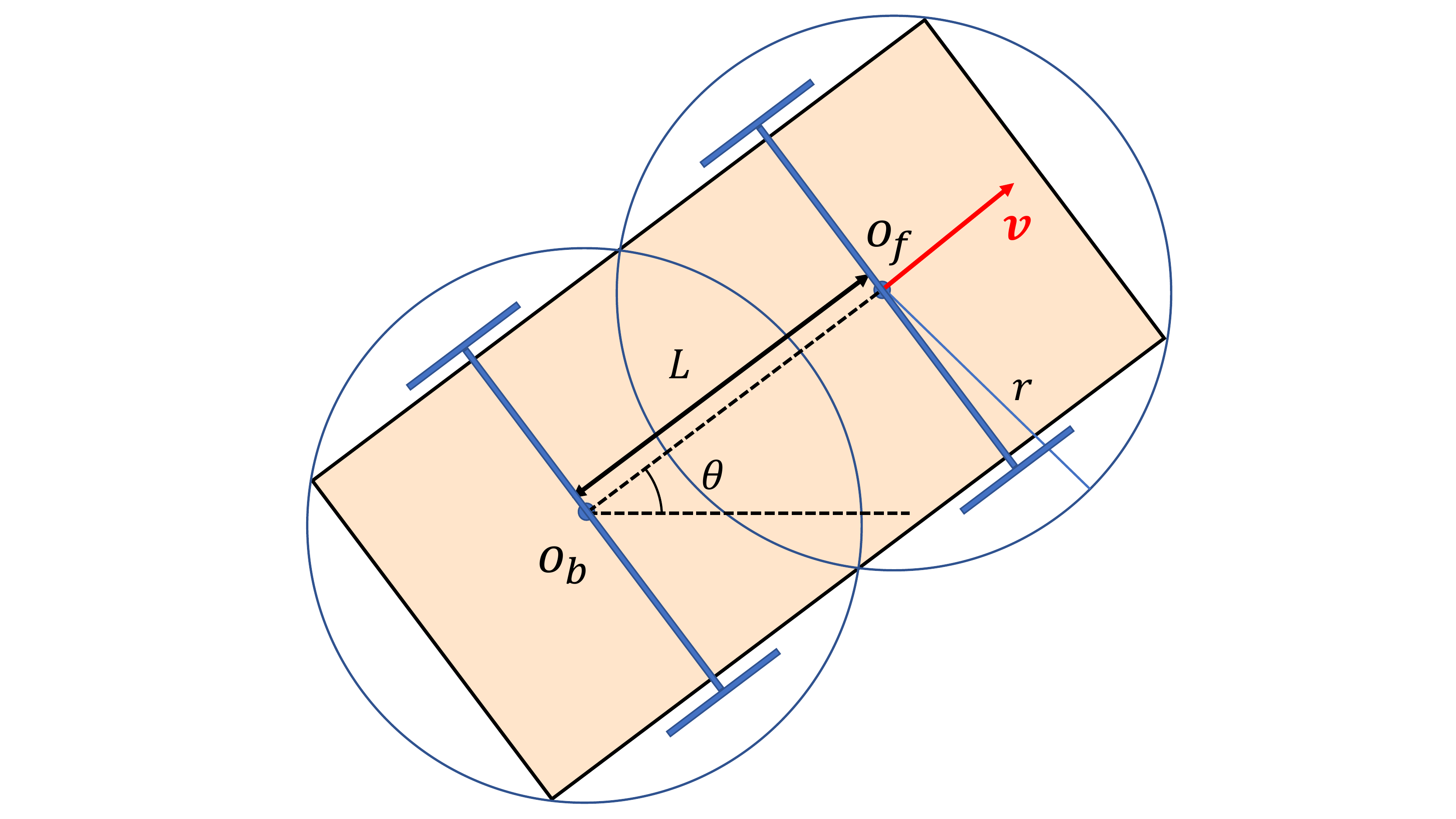}
    \caption{Two circles of radius $r$ are used to cover the whole robot. $\mathbf{o}_b$ is at the position of robot $\mathbf{p}$, and $\mathbf{o}_f$ is at $L$ along the robot's forward direction. The $\theta$ is the orientation angle of the robot. \label{fig:circle}}
\end{figure}

\subsubsection{Curvature Limit}
Nonholonomic robots always have curvature limitations due to their own factors, such as the steering angle of wheeled robots and roll angle of spherical robots\cite{zhang2021improved}. The curvature limitations can be written as:
\begin{equation}\label{eq:curvature_limit}
    \begin{aligned}
        &\frac{\vert \omega_k \vert}{\vert v_k \vert} \leq \kappa_{max}\\
        \forall &k\in\{0,1,2,...,n\},
    \end{aligned}
\end{equation}
where $\kappa_{max}$ is the maximum curvature of the nonholonomic robots. Considering that both the linear and angular velocity of the robot can be negative, we take absolute values for them.

\subsection{Safety Constraints}\label{sec:SafetyConstraints}

As shown in Fig.\ref{fig:trajectory}, we use the safe flight corridor construction algorithm in \cite{liu2017planning} to extract the free space in environments to generate an x-y 2D safe corridor composed of $\zeta$ H-represented convex polygons. A convex polygon with $m$ half-spaces is defined as:
\begin{equation}\label{eq:H_representation}
    \begin{aligned}
        \mathcal{C} = \bigcap_{j=1}^m H_j=\{ \mathbf{x}\in \mathbb{R}^2 | \mathbf{A}\mathbf{x} \leq \mathbf{b}\},
    \end{aligned}
\end{equation}
where $\mathbf{A}\in\mathbb{R}^{m\times2}$ and $\mathbf{b}\in\mathbb{R}^{m}$ are the descriptors of these $m$ half-spaces. The intersection of these half-spaces forms a polygon. To ensure that the robot will not collide with obstacles in the environment, we use two circles of radius $r$ to cover the entire robot, one of the circle centers is located at the robot position $\mathbf{p}$ as shown in Fig.\ref{fig:circle}.
We inflate the obstacles with radius $r$ before constructing the safe corridor. Therefore, the robot can be represented by two points:
\begin{equation}
    \begin{aligned}
        \mathbf{o}_{bi}&=\mathbf{p}_i \\
        \mathbf{o}_{fi}&=\mathbf{p}_i+L\begin{pmatrix}
            \cos{\theta}_i \\ \sin{\theta}_i
        \end{pmatrix}
    \end{aligned}
\end{equation}

\begin{figure}[t]
    \centering
    \subfigure[No additional restriction on the last state $P_{\sigma_{in_i}}$]{
        \label{fig:safety_a}
        \includegraphics[width=8cm]{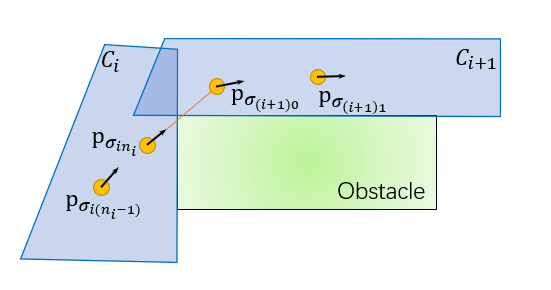}
    }
    \subfigure[Restrict the last state to the junction of two safe areas]{
        \label{fig:safety_b}
        \includegraphics[width=8.2cm]{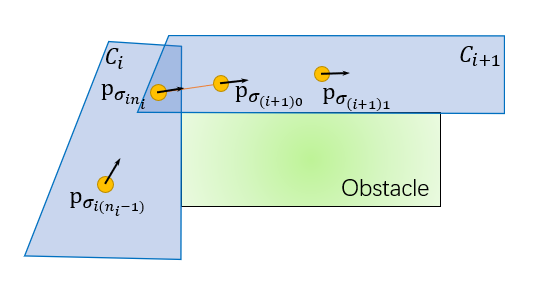}
    }
    \caption{The two figures show why we need an additional restriction on the last state of each polygon. As shown in Fig.\ref{fig:safety_a}, the last state of $\mathcal{C}_i$ and the first state of $\mathcal{C}_{i+1}$ are both in their corresponding safe zone, while the connecting line may collide with the obstacle. When the last state $P_{\sigma_{in_i}}$ is confined within the junction, as shown in Fig.\ref{fig:safety_b}, we can guarantee the robot will not collide with the obstacle.}
    \label{fig:safety}
\end{figure}

When constructing a corridor, if polygon $\mathcal{C}_i$ has $n_i$ initial path points inside, then we assign $n_i$ trajectory states in it, and use $\sigma_{ik}$ to denote the subscript of $k_{th}$ state within polygon $\mathcal{C}_i$. Then the safety constraints can be formulated as the points pair from $\sigma_{i0}$ to $\sigma_{in_i}$ are supposed to be inside the corresponding polygon $\mathcal{C}_i$:
\begin{equation}
    \begin{aligned}
        &\mathbf{A}_i\mathbf{o}_{kj}-\mathbf{b}_i \leq 0\\
        \forall i\in\{0,1,2,...,&\zeta\}, \forall j\in\{\sigma_{i0},\sigma_{i1},..., \sigma_{in_i}\}, \forall k\in\{b,f\}
    \end{aligned}
\end{equation}

For discrete states, the situation as shown in Fig.\ref{fig:safety_a} may occur. Although two points belonging to different polygons are both within their safe zone, the connecting line may collide with obstacles. To solve this problem, we place an additional constraint on the last state of each polygon, ensuring that it is at the junction of that polygon and the next polygon(except for the last polygon). Therefore, the constraint for the last state of each polygon is formulated as:
\begin{equation}
    \begin{aligned}
        \mathbf{p}_{\sigma_{in_i}}&\in\mathcal{C}_i\cap\mathcal{C}_{i+1}
        \Longleftrightarrow\\
        \begin{pmatrix}
            \mathbf{A}_i \\\mathbf{A}_{i+1}
        \end{pmatrix} &\mathbf{o}_{k\sigma_{in_i}}
        -
        \begin{pmatrix}
            \mathbf{b}_i \\ \mathbf{b}_{i+1}
        \end{pmatrix}
        \leq 0,
        \forall k\in\{b,f\}
    \end{aligned}
\end{equation}

\begin{figure}[t]
    \centering
    \subfigure[Without the gear shifting position constraint]{
        \label{fig:gear_shift1}
        \includegraphics[width=8cm]{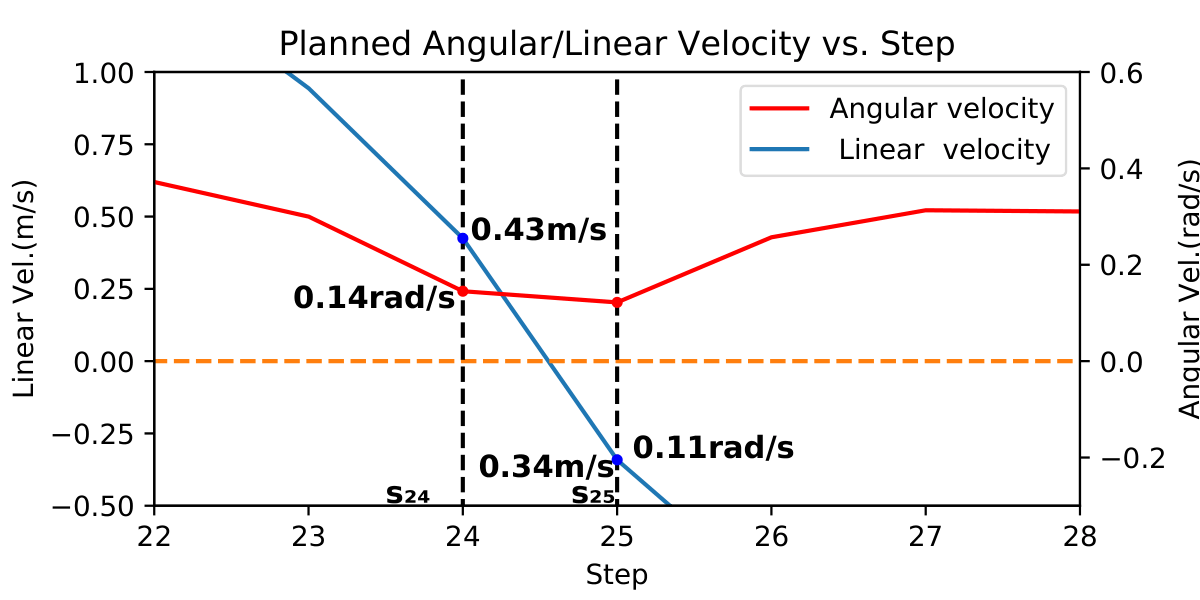}
    }
    \subfigure[With the gear shifting position constraint]{
        \label{fig:gear_shift2}
        \includegraphics[width=8cm]{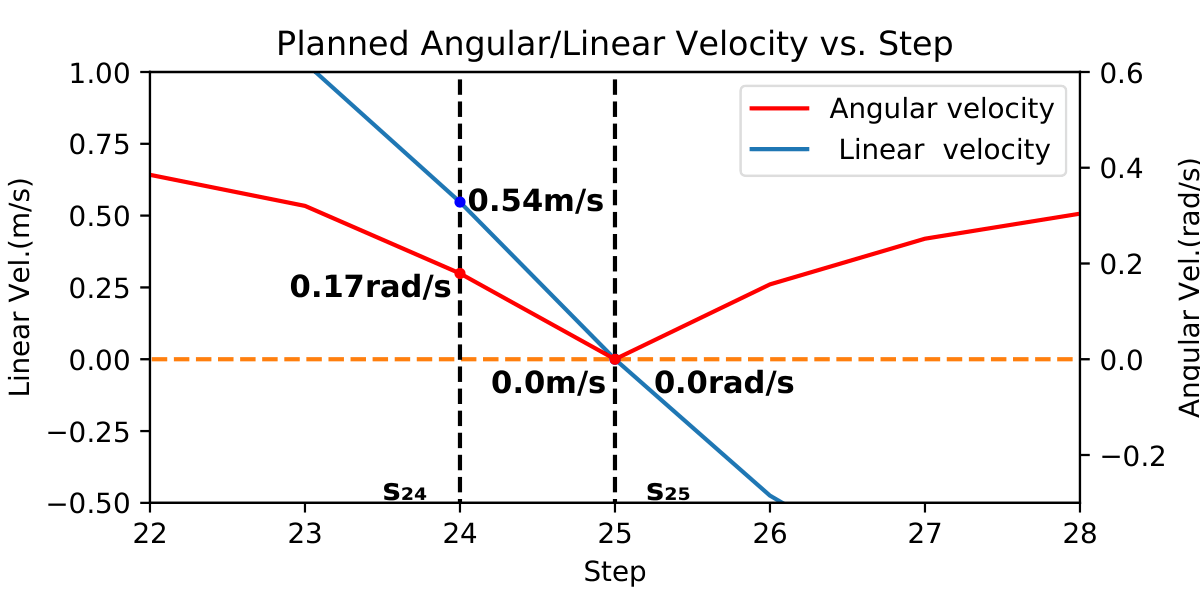}
    }
    \caption{Comparison of whether to add gear shifting position constraints.}
    \label{fig:gear_shift}
\end{figure}

\subsection{Gear Shifting Position Constraint}
The above formulation is able to apply to forward and backward situation. However, the speed may do not reach zero at a specific state as shown in Fig.\ref{fig:gear_shift}. In Fig.\ref{fig:gear_shift1}, even though the states in step 24 and 25 meet the curvature constraint, the velocity changes from positive to negative without zero point, and the curvature constraint will be violated during this process. Therefore, We have to ensure that the speed of the robot to be zero at the shifting position:
\begin{equation}\label{eq:gear_shift_limit}
    v_iv_{i+1}\geq0,
    \forall i\in\{0,1,2,...,n-1\}
\end{equation}

As we can see, the linear and angular velocities at step25 are 0 in Fig.\ref{fig:gear_shift2} after adding the gear shifting position constraint. Furthermore, the reason why we do not get the shifting point from the initial path and then limit the speed is that we want the optimization program determine whether and where to shift gears.

\section{Discrete States-Based Trajectory Optimization}
\subsection{Problem Formulation}
In Sect.\ref{sec:Discrete_States}, we discuss about the discrete states of nonholonomic robots. Since trajectory optimization for nonholonomic robots is a difficult problem due to its severe non-convexity, we formulate the trajectory optimization problem as a problem with several penalty terms and box constraints:
\begin{equation}\label{eq:problem}
    \begin{aligned}
        \mathop{\min}\quad &J = \lambda_oJ_{o} + \lambda_{eq}J_{eq} + \lambda_{ie}J_{ie}\\
        \text{s.t.}\quad &v_{min} \leq v_k \leq v_{max}\\
        &a_{min} \leq a_k \leq a_{max}\\
         &\qquad\quad\ \ t_k > 0\\
        \forall &k\in \{0,1,2,...,n\},
    \end{aligned}
\end{equation}
where $J_o$ is the optimization objective function, $J_{eq}$ is for equality constraints, and $J_{ie}$ is for inequality constraints. $\lambda_o, \lambda_{eq}, \lambda_{ie}$ are the weights for each penalty terms.

\subsection{Optimization Objective Function}\label{sec:OptimizationGoal}
We always want the robots to get the destination smoothly and quickly. Generally speaking, the smoothness of trajectories depends on linear and angular aspects. Therefore, we formulate the smoothness optimization function with the jerk and angular acceleration. The jerk $j$ and angular acceleration $\alpha$ from $t_k$ to $t_{k+1}$ can be written as:
\begin{equation}
    \begin{aligned}
       j_k &= \frac{a_{k+1}-a_k}{t_k}\\
       \alpha_k &= \frac{\omega_{k+1}-\omega_k}{t_k}
    \end{aligned}
\end{equation}

Then the optimization goal $J_O$ is the integral of the square of jerk and angular acceleration. The time function is formulated as the sum of all time intervals in order to make the robot complete the moving task in a shorter time:

\begin{equation}
    \begin{aligned}
        J_o &= \sum_{i=0}^{n-1} (\int_{t_i}^{t_{i+1}}j_i^2dt + \int_{t_i}^{t_{i+1}}\alpha_i^2dt +\delta_tt_i)\\
        &= \sum_{i=0}^{n-1} (\frac{(a_{i+1}-a_i)^2}{t_i} + \frac{(\omega_{i+1}-\omega_i)^2}{t_i} +\delta_tt_i),
    \end{aligned}
\end{equation}
where $\delta_t$ is the time weighting parameter.

\subsection{Equality Constraints}
The equality constraints in the problem comes from the redundant states relationship and waypoints constraints. We first rewrite (\ref{eq:relationship}) for brevity:
\begin{equation}
    \begin{aligned}
        &\mathbf{s}_{k+1}\triangleq f(\mathbf{s}_k,\mathbf{c}_k,t_k)\\
        \forall k&\in\{0,1,2,...,n-1\}.
    \end{aligned}
\end{equation}

Then we express the equality constraint in terms of a quadratic penalty:
\begin{equation}
    \begin{aligned}
       J_{eq}=&\sum_{i=0}^{n-1}\Vert\mathbf{s}_{i+1}-f(\mathbf{s}_i,\mathbf{c}_i,t_i)\Vert^2+\Vert\mathbf{s}_0-\mathbf{s}_s\Vert^2\\
       &+\Vert\mathbf{s}_n-\mathbf{s}_g\Vert^2
    \end{aligned}
\end{equation}

\subsection{Inequality Constraints}
Although box constraints of limited velocity and acceleration are inequality constraints, we have simpler ways to restrict them than writing in $J_{ie}$. We will discuss them later, so the box constraints will not include here.

Other inequality constraints are formulated as:
\begin{equation}
    \begin{aligned}
        J_{ie}=&\sum_{i=0}^n\delta_\kappa F_\kappa(v_i,\omega_i)
        +\sum_{i=0}^\zeta\delta_s F_s(\mathbf{A}_i,\mathbf{b}_i)\\
        +&\sum_{i=0}^{n-1}\delta_v F_v(v_i,v_{i+1}),
    \end{aligned}
\end{equation}
where $F_\kappa, F_s, F_v$ are for the curvature constraint, safety constraint and gear shifting position constraint, respectively. And $\delta_\kappa, \delta_s$ are weights for each terms.
\begin{equation}
    \begin{aligned}
        F_\kappa(v_i,\omega_i)&=L(\omega_i^2  - v_i^2\kappa_{max}^2)\\
        F_s(\mathbf{A}_i,\mathbf{b}_i)&= \sum_{j=0}^{n_i}L(\mathbf{A}_i\mathbf{p}_{ij}-\mathbf{b}_i)+L(\mathbf{A}_{i+1}\mathbf{p}_{in_i}-\mathbf{b}_{i+1})\\
        F_v(v_i,v_{i+1})&=L(-v_iv_{i+1})
    \end{aligned}
\end{equation}

We reformulate the curvature constraint (\ref{eq:curvature_limit}) due to its discontinuity at $v=0$, and $L(\cdot)$ is a twice continuously differentiable penalty function.
\begin{equation}
    \begin{aligned}
        L(x)=\begin{cases}
            0 &(x<=0)\\
            x^3 &(0<x<x_j)\\
            3x_jx^2-3x_j^2x+x_j^3 &(x\geq x_j),
        \end{cases}
    \end{aligned}
\end{equation}
where $x_j$ is the demarcation point of the quadratic penalty and the cubic penalty.

\subsection{Nonlinear Optimization with Box Constraints}
The formulated problem in (\ref{eq:problem}) has many quadratic terms. Therefore, we choose quasi-Newton methods to solve it. L-BFGS-B was proposed by Byrd et al.\cite{L-BFGS-B} for solving large nonlinear optimization problems with simple bounds as:
\begin{equation}\label{eq:LBFGS-B_problem}
    \begin{aligned}
        \min\quad &h(x)\\
        \text{s.t.}\quad &l \leq x \leq u
    \end{aligned}
\end{equation}

Here we explain the algorithm briefly. For the problem in (\ref{eq:LBFGS-B_problem}), at each iteration,
a quadratic model of $h$ at current $x_k$ is formed as
\begin{equation}
    m_k(x)=h(x_k)+g_k^T(x-x_k)+\frac{1}{2}(x-x_k)^TB_k(x-x_k),
\end{equation}
where $g_k^T$ and $B_k$ represent the gradient vector and the positive definite limited-memory Hessian approximated by L-BFGS\cite{LBFGS}, respectively.

The generalized Cauchy point $x^c$ in the feasible region will be computed, which is defined as the first local minimizer of $m_k(x)$ along the negative gradient direction. The variables whose value at $x^c$ is at lower or upper bound comprise the active set $\mathcal{A}(x^c)$, and minimize $m_k(x)$ by the other variables as an unconstrained problem.
\begin{equation}\label{eq:unconstrainted_problem}
    \min \{m_k(x):x_i=x_i^c,\forall i\in \mathcal{A}(x^c)\}
\end{equation}

\begin{figure}[tb]
    \centering
    \subfigure[Maze]{
        \label{fig:maze}
        \includegraphics[width=8.5cm]{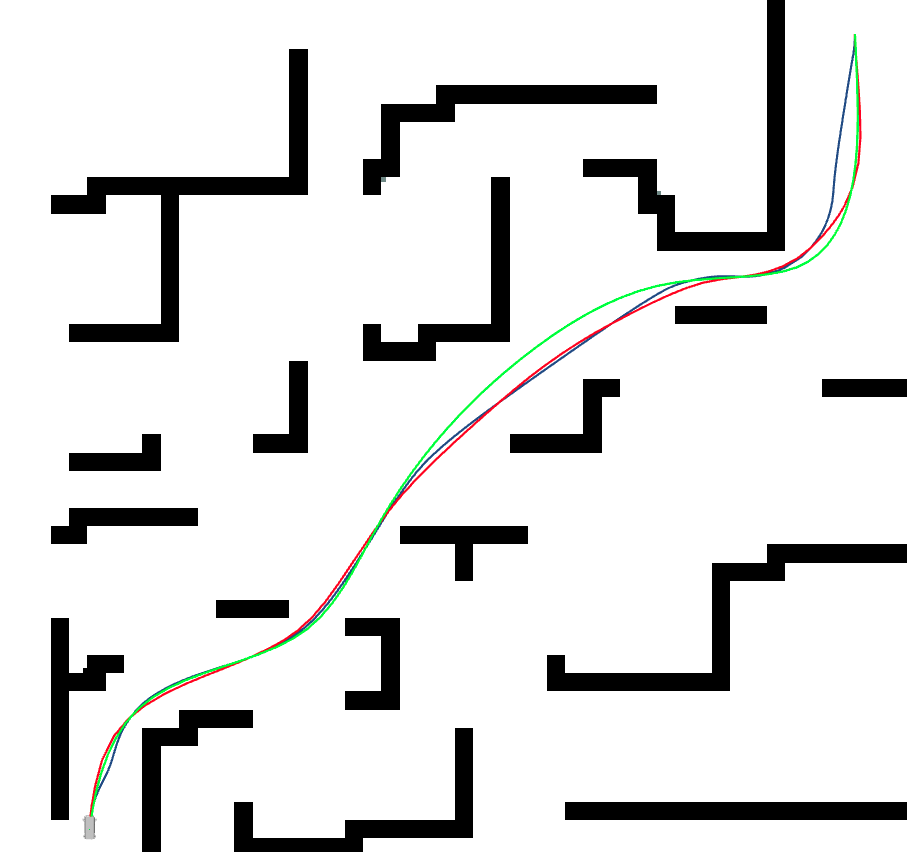}
    }
    \subfigure[Parking lot]{
        \label{parking_lot}
        \includegraphics[width=8.5cm]{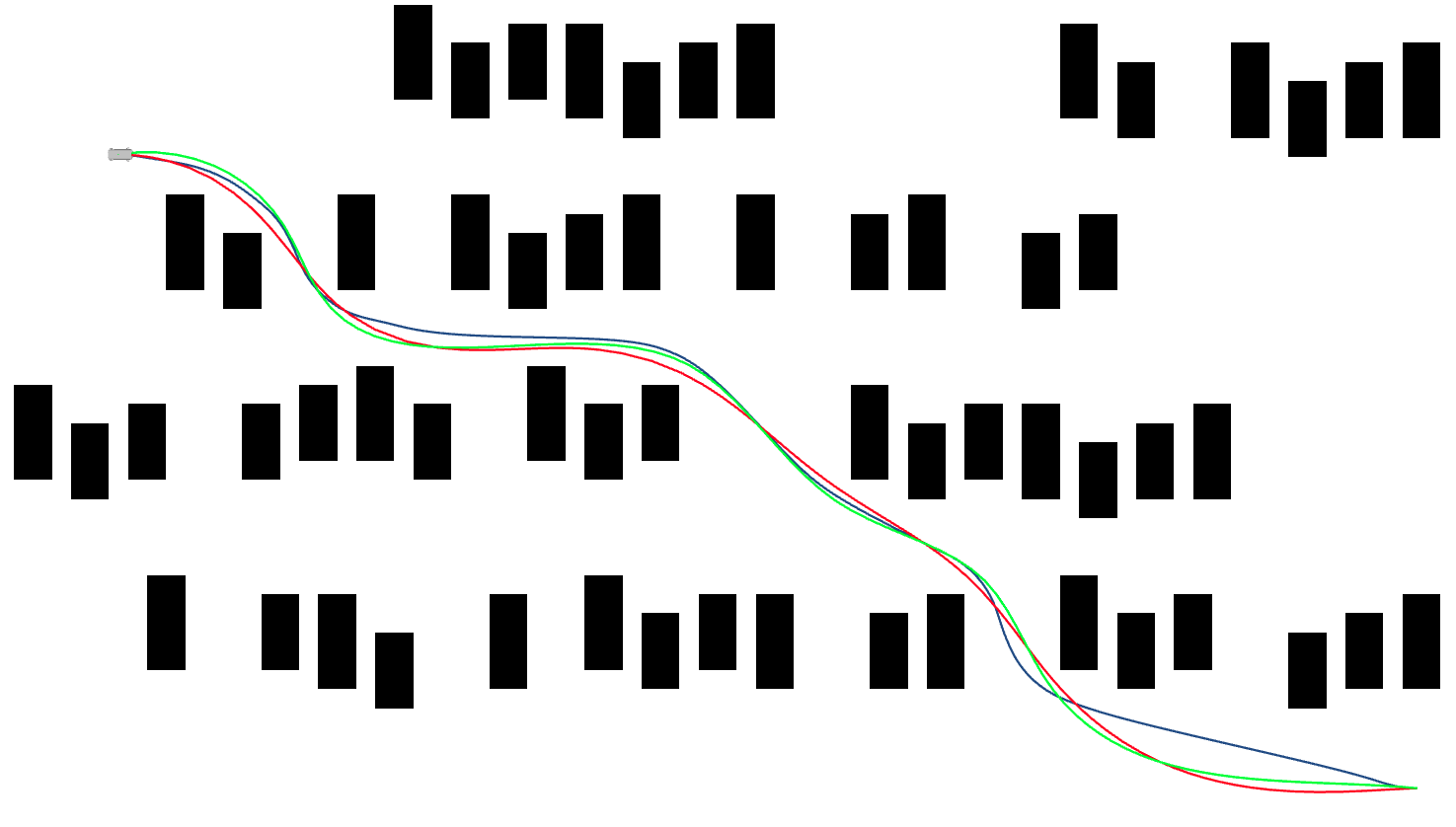}
    }
    \subfigure[Irregular map]{
        \label{s_bend}
        \includegraphics[width=8.5cm]{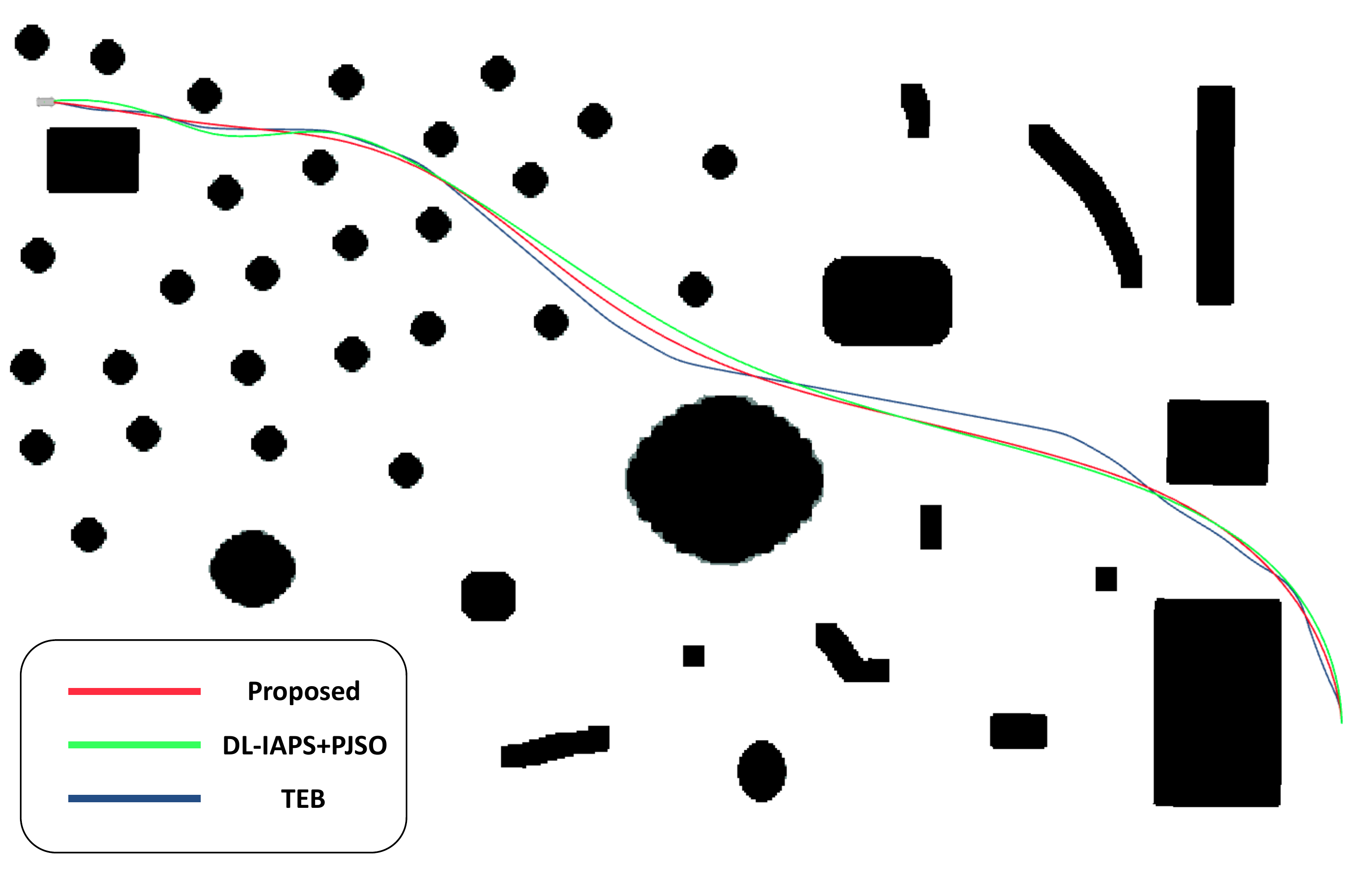}
    }
    \caption{Trajectory visualization in simulation.}
    \label{fig:simulation}
\end{figure}

The bound constraints for the other variables are satisfied by truncating the path toward the solution of (\ref{eq:unconstrainted_problem}) to obtain the approximate solution $\Bar{x}_{k+1}$.
After that, the line search direction is determined as $d_k=\Bar{x}_{k+1}-x_k$ and the steplength will be calculated by strong Wolfe condition.

To speed up the L-BFGS-B, we use the result of each solution as the initial guess to warm start the next solution, which can help us greatly reduce the optimization time during replanning.

\begin{figure}
    \includegraphics[width=8.5cm]{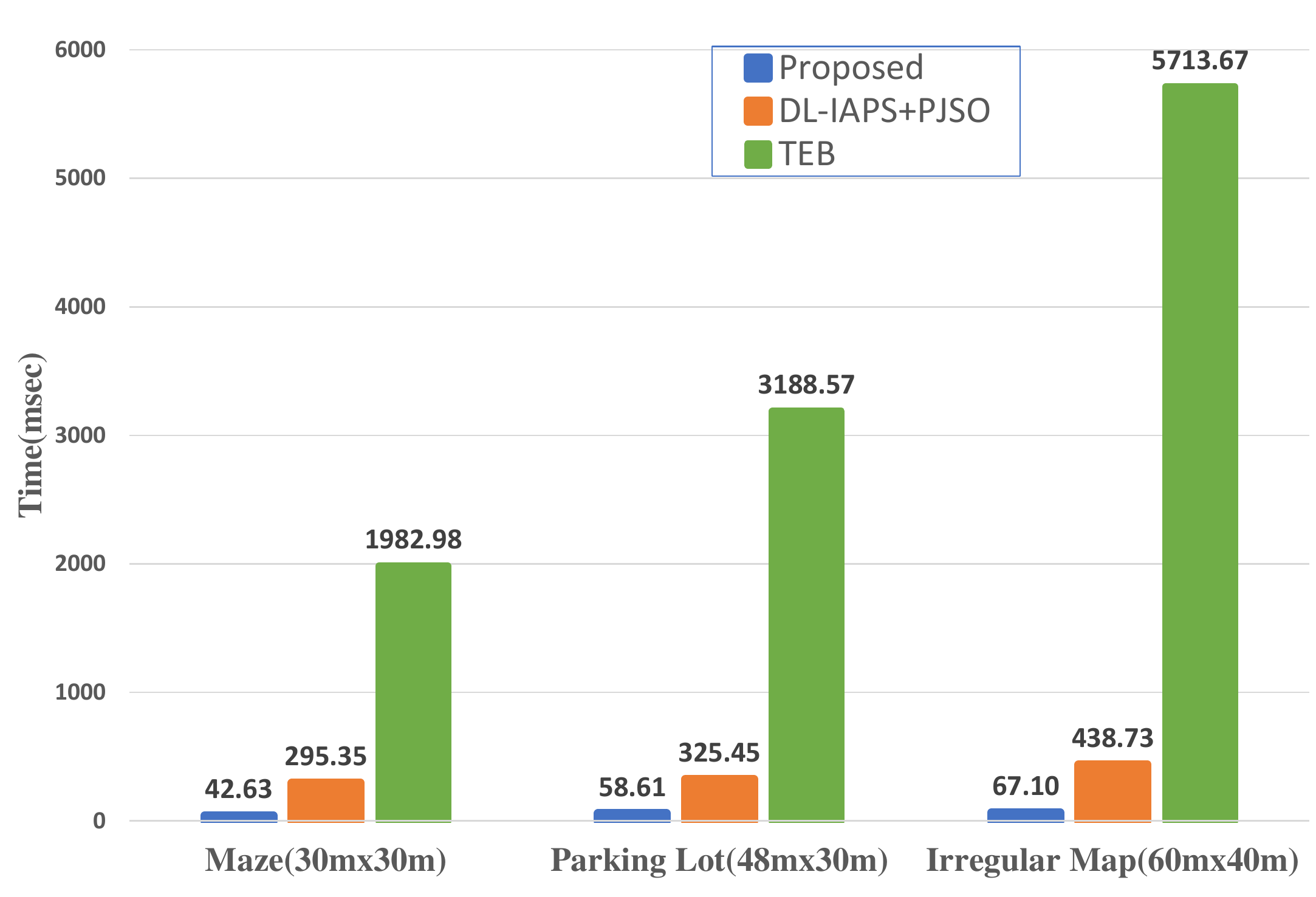}
    \caption{\centering{Average computation time comparison of proposed DSTP-planner against DL-IAPS+PJSO and TEB.}}
    \label{fig:time}
\end{figure}

\renewcommand\arraystretch{1.5}
\begin{table}[t] \label{tab:simulation}
\centering
{
\caption{Performance comparison in different scenarios}
    \label{tab:simulation}
}
\resizebox{\linewidth}{!}{\begin{tabular}{c|c|c|c|c|c|c|c}
\hline
\multirow{2}{*}{Scenario} & \multirow{2}{*}{Method} & \multicolumn{2}{c|}{\begin{tabular}[c]{@{}c@{}}Velocity\\ ($m/s$)\end{tabular}} & \multicolumn{2}{c|}{\begin{tabular}[c]{@{}c@{}}Acceleration\\ ($m/s^2$)\end{tabular}} & \multicolumn{2}{c}{\begin{tabular}[c]{@{}c@{}}Jerk\\ ($m/s^3$)\end{tabular}}
\\ \cline{3-8}  
 & & mean  & max & mean    & max     & mean   & max
\\ \hline \hline
\multirow{3}{*}{\begin{tabular}[c]{@{}c@{}}Maze\\ (30mx30m)\end{tabular}} & DSTP & 2.27  & 3.00   & \textbf{0.27}     & \textbf{0.92}    & \textbf{0.11} & \textbf{0.23} 
\\ \cline{2-8} 
& DL-IAPS+PJSO & 1.83 & 3.00   & 0.32  & 2.00  & 0.24   & 1.00          
\\ \cline{2-8} 
 & TEB & \textbf{2.67} & 3.00 & 0.41  & 2.00    & 2.04   & 14.87         
 \\ \hline \hline
\multirow{3}{*}{\begin{tabular}[c]{@{}c@{}}Parking Lot\\ (48mx30m)\end{tabular}}   & DSTP          & 2.42  & 3.00  & \textbf{0.21}     & \textbf{0.93}    &\textbf{0.09} & \textbf{0.21} 
\\ \cline{2-8} 
 & DL-IAPS+PJSO & 1.81  & 3.00  &0.24    & 2.00   & 0.20    & 1.00          
 \\ \cline{2-8} 
 & TEB & \textbf{2.71} & 3.00 & 0.72     & 2.00  & 4.22 & 15.64         
 \\ \hline \hline
\multirow{3}{*}{\begin{tabular}[c]{@{}c@{}}Irregular Map\\ (60mx40m)\end{tabular}} & DSTP          & 2.57  & 3.00  & \textbf{0.19}     & \textbf{0.94}    &\textbf{0.08} & \textbf{0.20} 
\\ \cline{2-8} 
 & DL-IAPS+PJSO & 1.88  & 3.00  &0.24 & 2.00    & 0.19    & 1.00  
 \\ \cline{2-8} 
 & TEB & \textbf{2.65} & 3.00 & 0.88    & 2.00     & 4.80    & 18.62       \\ \hline
\end{tabular}}
\end{table}

\begin{figure*}
    \centering
    \includegraphics[width=18cm,height=8cm]{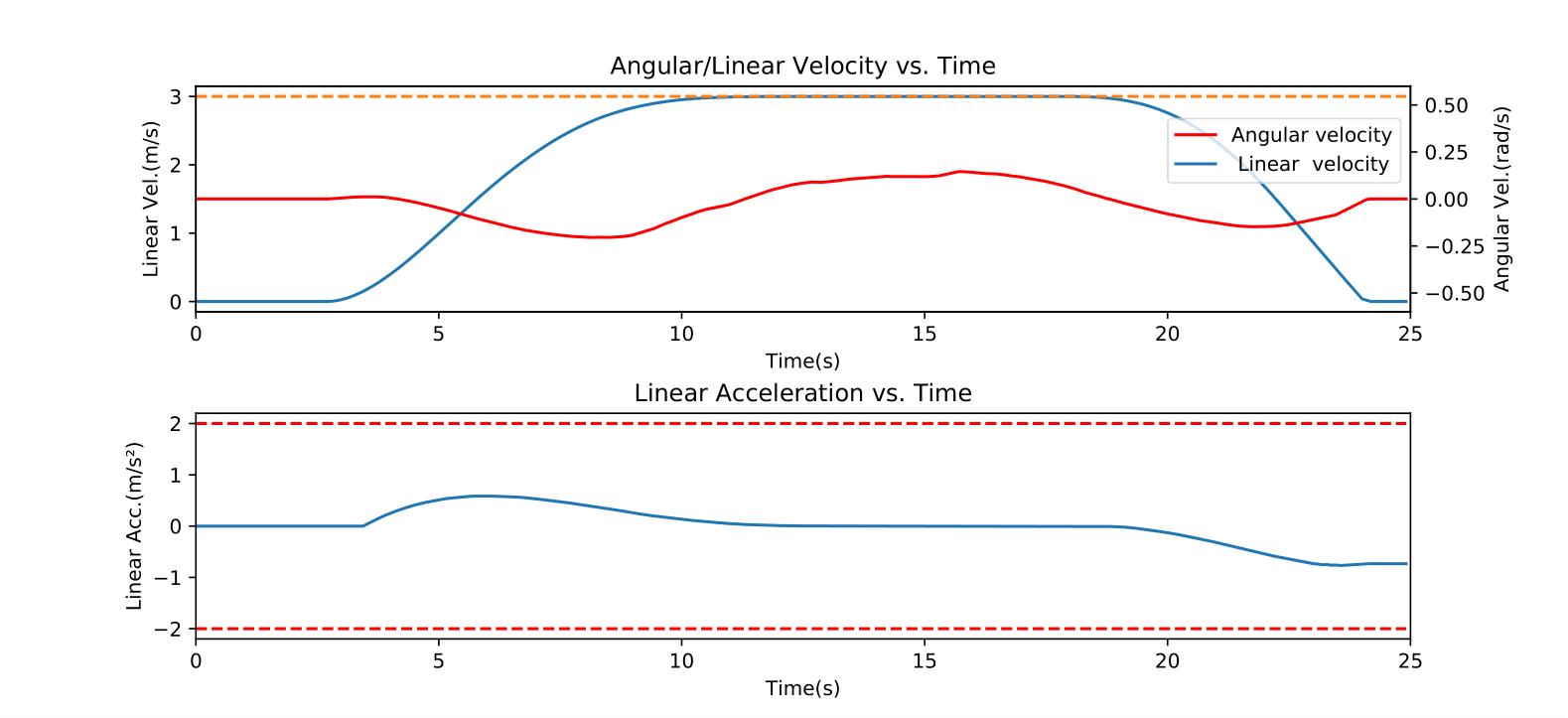}
    \caption{\centering{Visualization of DSTP trajectory planning in irregular map with linear/angular velocity and linear acceleration.}}
    \label{fig:curve}
\end{figure*}

\section{Experiment Results}
To evaluate the performance of the DSTP algorithm, we conducted simulation experiments and real-world tests with a car-like robot. 

\begin{figure}
    \centering
    \includegraphics[width=8cm]{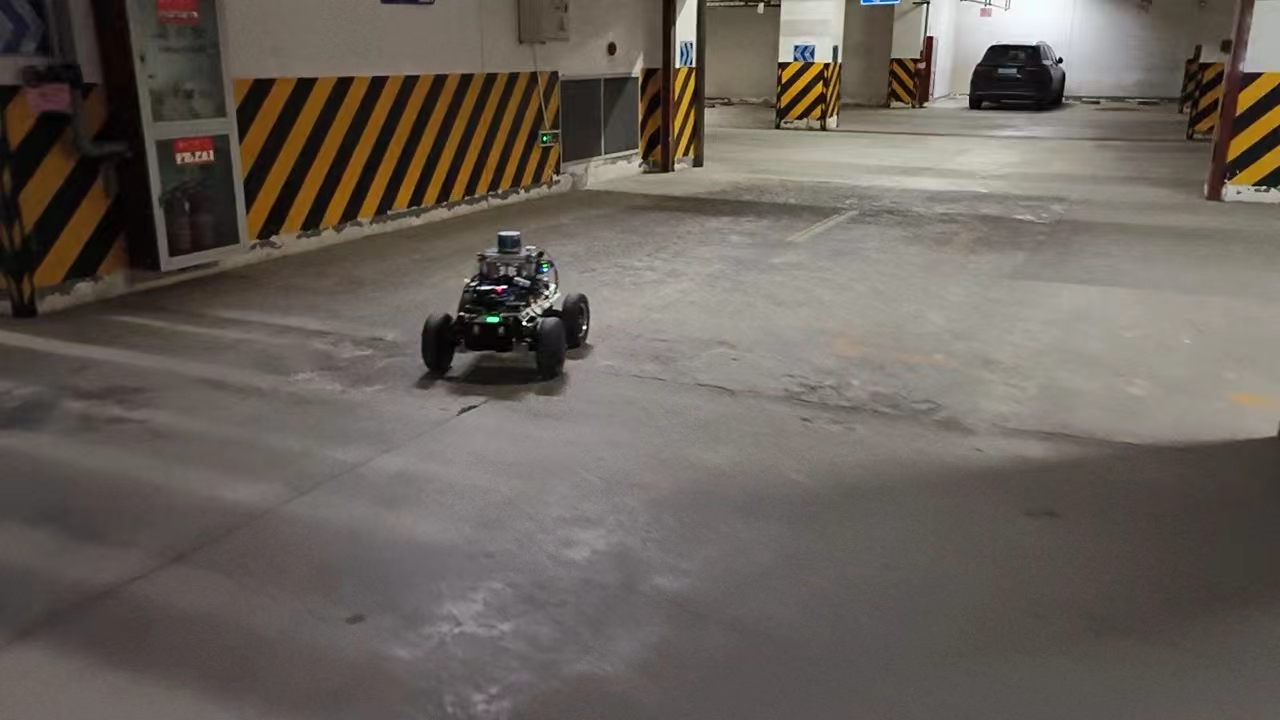}
    \caption{\centering{Real-world experiments with the car-like robot.}}
    \label{fig:car}
\end{figure}

\subsection{Simulation Experiments}
We compare the proposed algorithm with TEB\cite{TEB2} and DL-IAPS+PJSO\cite{PJSO} in three different scenarios: maze map($30m \times 30m$), parking lot($48m \times 30m$) and an irregular map with many obstacles($60m \times 40m$). Three planners all run twenty times in each scenario. The maximum velocity and acceleration are set to $3m/s$ and $2m/s^2$, respectively. A head-only L-BFGS-B library\footnote{https://github.com/yixuan/LBFGSpp} is used for DSTP. Hybrid A*\cite{dolgov2010path} algorithm is used for three planners as the front-end path searching to provide the initial path.

\begin{figure}
    \centering
    \includegraphics[width=8cm]{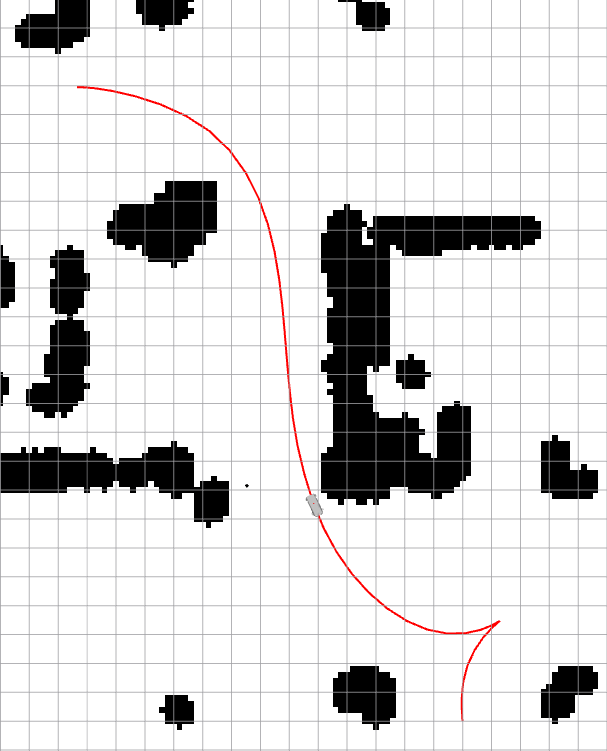}
    \caption{{Trajectory generated by DSTP of a real-world experiment, where red curve is the trajectory followed by the robot and the cell size is set to $1m \times 1m$.}}
    \label{fig:real-world}
\end{figure}

Trajectories generated by three planners are illustrated in Fig.\ref{fig:simulation}. In order to better compare the efficiency of the three planners, we draw the average computation time in each scenario in Fig.\ref{fig:time}. It should be noted that the time of DSTP is the time of the first solution, excluding the time of replanning using the previous solution as the warm start for fairness. The histogram shows that our algorithm has an obvious advantage in efficiency compared with the other two planners. Even in the large and irregular random map, the DSTP can complete the planning in around 70 msec.

The performance statistics including mean and max computation time, velocity, acceleration, and jerk are shown in Tab.\ref{tab:simulation}. In the three scenarios, the maximum speed and acceleration of the three planners can be all well limited. The average speed of TEB is the fastest, followed by DSTP. However, due to the lack of consideration of trajectory smoothing, the average acceleration and jerk of TEB are far greater than those of PJSO and DSTP. The average acceleration and jerk of DSTP are the smallest, which shows that our planner has the lowest control effort and the smoothest trajectories.
Besides, Fig.\ref{fig:curve} shows the dynamic profile of the trajectory planned by the proposed algorithm. The dashed lines denote the limit of maximum speed and acceleration.

\subsection{Real-World Experiments}
To better verify the feasibility and performance of our algorithm in the real world, we conducted experiments with a car-like robot in an underground garage, as shown in Fig.\ref{fig:car}. The robot starts from an initial state, crosses the garage, and finally reverses to the goal state. The trajectory executed by the robot is shown in Fig.\ref{fig:real-world}. Obstacle detection and robot localization are provided by the laser radar installed on the top of the robot. The maximum velocity and acceleration are set to $2m/s$ and $2m/s^2$, respectively.
Statistics in real-world experiments are quantified in Tab.\ref{tab:real-world}. In the $21m \times 26m$ garage, the computation time of DSTP is around 30 msec, which is acceptable for real-time application. In addition, the robot can maintain a high speed throughout the whole moving process without exceeding the maximum speed and acceleration limits, and ensure smooth movement through a low jerk.

\begin{table}[]
\centering
{
\caption{Statistics in Real-world Experiments}
    \label{tab:real-world}
}
\begin{tabular}{ccccc}
\hline
 &
  \begin{tabular}[c]{@{}c@{}}Computation Time\\ ($msec$)\end{tabular} &
  \begin{tabular}[c]{@{}c@{}}Velocity\\ ($m/s$)\end{tabular} &
  \begin{tabular}[c]{@{}c@{}}Acceleration\\ ($m/s^2$)\end{tabular} &
  \begin{tabular}[c]{@{}c@{}}Jerk\\ ($m/s^3$)\end{tabular} \\ \hline
mean &
  29.33 &
  1.36 &
  0.25 &
  0.13 \\ \hline
max &
  35.42 &
  2.00 &
  0.64 &
  0.42 \\ \hline
\end{tabular}
\end{table}

\section{Conclusion}
In this letter, we present a novel discrete states-based trajectory optimization algorithm. The proposed algorithm uses the discrete states of the robot to carry out trajectory optimization with kinematics and dynamics constraints. Through many simulation experiments compared with the other two planners, we have proved the advantages in efficiency and trajectory quality of our algorithm. Real-world experiments are also conducted to validate that our algorithm is efficient and can maintain a high speed with a smooth trajectory and low control effort without exceeding the limit velocity and acceleration.
We plan to extend our work to more applications including 1) autonomous driving in unstructured and structured environments and 2) highly dynamic environments with many other fast-moving objects.

The proposed method still has some flaws, for example, the relationships between states are still approximate, which will bring errors and is not conducive to trajectory tracking. We will work on more precise and concise optimization problem formulations in the future.

\ifCLASSOPTIONcaptionsoff
  \newpage
\fi

\bibliographystyle{ieeetr} 
 
\bibliography{bibtex/cite.bib}    

\end{document}